\documentclass[10pt,english]{article}
\usepackage[T1]{fontenc}
\usepackage{babel}
\usepackage{amsmath}
\usepackage{amsthm}
\usepackage{graphicx}
\usepackage[authoryear]{natbib}
\usepackage[compact]{titlesec}
\usepackage{ifthen}
\usepackage{wrapfig}
\newboolean{submit}
\setboolean{submit}{false}
\usepackage{hyperref}

\makeatletter
\theoremstyle{plain}

\ifx\proof\undefined
\newenvironment{proof}[1][\protect\proofname]{\par
	\normalfont\topsep6\p@\@plus6\p@\relax
	\trivlist
	\itemindent\parindent
	\item[\hskip\labelsep\scshape #1]\ignorespaces
}{%
	\endtrivlist\@endpefalse
}
\providecommand{\proofname}{Proof}
\fi

\ifthenelse{\boolean{submit}}
{\usepackage{neurips_2019}}
{\usepackage[preprint]{neurips_2019}}
 
 \usepackage{algorithm}
\usepackage{algorithmic}

\title{Differential Description Length for Model Selection in Machine Learning}

\author{Mojtaba Abolfazli, Anders H{\o}st-Madsen\thanks{Also affiliated with Shenzhen Research Institute of Big Data, CUHKSZ, as a visiting professor}, June Zhang \\
Department of Electrical Engineering \\
University of Hawaii at Manoa \\
\{mojtaba, ahm, zjz\}@hawaii.edu}

\makeatother

\providecommand{\theoremname}{Theorem}

\begin{document}
\title{Differential Description Length for Hyperparameter Selection in Machine Learning}
\maketitle

\begin{abstract}
This paper introduces a new method for model selection
and more generally hyperparameter selection in machine
learning. Minimum description length (MDL) is an established
method for model selection, which is however not directly
aimed at minimizing generalization error, which is often the primary goal
in machine learning.
The paper demonstrates a relationship
between generalization error and a difference of description lengths of the training data; we call
this difference differential description length (DDL). This allows
prediction of generalization error from the training data
\emph{alone} by performing encoding of the training data. 
DDL can then be used for model selection by choosing the model with the smallest predicted generalization error.
We show how this method can be used for linear
regression
%, logistic regression, 
and neural networks and deep learning.
Experimental results show that DDL leads to smaller generalization error than cross-validation and
traditional MDL and Bayes methods.
\end{abstract}

\section{Introduction}

Minimum description length (MDL) is an established method for model
selection. It was developed in the pioneering papers by  \citet{Rissanen78,Rissanen83,Rissanen86},
and has found wide use \citep{GrunwaldBook}.

In this paper we consider description length 
%(which we use as wider term
%than MDL) 
for machine learning problems.
Specifically, we consider a supervised learning problem with features $x$ and labels
$y$. Given a set of
training data $((x_1,y_1),\ldots,(x_n,y_n))$ we want to find a predictor $f(x;\theta_{h},h)$
of $y$. Here $\theta_{h}$ is a a set of parameters that are estimated
from the training data, and $h$ is a set of \emph{hyperparameters} that are chosen;
these are typically the model order, e.g., number of hidden units
and layers in neural networks, but also quantities like  regularization parameters 
and early stopping times \citep{BishopBook}.
The goal of learning is to minimize the test error, or generalization error,
\begin{align}
	E_{x,y}[L(y,f(x;\theta_{h},h))] \label{eq:generror}
\end{align}
for some loss function $L$. Here $E_{x,y}$ makes explicit that the expectation is with
respect to $x,y$.
However, only the empirical loss (risk)
is available:
\begin{align}
	\frac 1 n \sum_{i=1}^{n}L(y_{i},f(x_{i};\theta_{h},h))] \label{eq:empirical}
\end{align}
Purely
minimizing the empirical loss with respect to the hyperparameters $h$ can lead to overfitting \citep{BishopBook,HastieBook}.
Description length is one method to avoid this.
Using MDL for model selection in
learning has been considered before, e.g., \citet{Grunwald11,Watanabe13,WatanabeRoos15,KawakitaTakeuchi16,Alabdulmohsin18}.

MDL aims to find the best model for fitting data according to an abstract criterion: which model gives the shortest codelength 
for data. When one of the models is the ``correct'' one, i.e., it has actually generated the data,
MDL might have good performance in terms of error probability in model selection.
On the other hand, in machine learning, the aim of model selection has  a concrete
target, namely minimizing (\ref{eq:generror}). Additionally, none
of the models might be the ``correct'' one. In this paper we show how MDL can
be modified so that it directly aims at minimizing (\ref{eq:generror}).

We measure performance by \emph{regret},
\begin{align}
	\text{Regret} = E_{x,y}[L(y,f(x;\theta_{h},\hat h))]
	- \inf_h E_{x,y}[L(y,f(x;\theta_{h},h))] \label{eq:regret}
\end{align}
where $\hat h$ is the chosen value of $h$.
%The (minimum) description length of data is the number of bits
%required to describe the data by a (universal) lossless source coder \citep{CoverBook}.
%This is the starting point of MDL, with much literature
% focused on finding (approximate) formulas for this codelength \citep{GrunwaldBook}, e.g., the well-known
%formula from \citet{Rissanen83}
%\begin{align*}
%  C = -\log f({\mathbf x};\hat\theta_{\text{MLE}})+\frac k 2\log n
%\end{align*}
%where $k$ is the dimension of $\theta$ and $n$ is the number of samples.
% In this paper
%we chart a different route. We find the description
%length by designing actual (sequential) source
%coders and applying these to data; this 
%follows the path of predictive MDL \citep{Rissanen86}, sequential NML \citep{RoosRissanen08}, and SSM \citep{Sabeti2017ISITpredictive}.
%We do not believe that simple approximate formulas can always
%capture the complexity of complex learning algorithms, an
%argument also used against MDL in for example
%\citet{BishopBook}. As an example, the impact
%of regularization parameters are not characterized by simply counting parameters.

\section{Theory}\label{sec:Theory}

We consider a supervised learning problem with features $x$ and labels
$y$. The data $(x,y)$ is governed by a probability law $P_{\theta}(y|x)p(x)$,
where $\theta$ is a parameter vector; these can be probability mass
functions or probability density functions. Notice that the distribution
of the features $x$ does not depend on $\theta$. 

We are given a training set $\{(x_{i},y_{i}),i=1,\ldots,n\}$ which
we assume is drawn iid from the distribution $P_{\theta}(y|x)p(x)$. We
use the notation $(x^{n},y^{n})$ to denote the whole training set; we might
consider the training set to be either fixed or a random vector.
The problem we consider is, based on the training data, to estimate the probability distribution  $P_{\hat\theta}(y|x)$ so as to minimize the log-loss or cross-entropy
\begin{equation}
E_{}\left[-\log P_{\hat \theta}(y|x)\right]. \label{LogLoss.eq}
\end{equation}
%The expectation here is over both test data and the training set, $\{(x,y);\{x^{n},t^{n})\}$,
%with respect to the distribution $P_{\theta}(y|x)p(x)$, for a \emph{fixed}
%$\theta$. %We will discuss other loss functions later.

\subsection{\label{SourceCoding.sec}Universal Source Coding and Learning}

In this section
we assume  that the data is from a finite alphabet. Based on the
training data $(x^{n},y^{n})$ we want to learn a good estimated probability
law $\hat{P}_L(y|x)$ (which need not be of the type $P_{\theta}(y|x)$),
and consider as in (\ref{LogLoss.eq}) the log-loss
\begin{equation}
C_{L}(n)=E_{x_0,y_0,x^n,y^n}\left[-\log\hat{P}_L(y_0|x_0)\right]\label{CL1.eq}
\end{equation}
Here $x_0,y_0$ is the test data.
Importantly, we can interpret $C_{L}(n) $ as a\emph{ codelength} as follows. By a codelength we mean the number of bits
required to represent the data without loss (as when zipping a file).
First the encoder is given the training data
$(x^{n},y^{n})$ from which it forms $\hat{P}_L(y|x)$; this is shared
with the decoder. Notice that this sharing is done ahead of time, and does not contribute to the codelength. Next, the encoder is given \emph{new data} $(\tilde x^{m},\tilde y^{m})$.
The decoder  knows $\tilde x^{m}$ but not $\tilde y^{m}$. The encoder encodes
$\tilde y^{m}$ using $\hat{P}_L(\tilde y_i|\tilde x_i)$ (using an algebraic coder \citep{CoverBook}),
and the decoder, knowing
$\tilde x^{m}$, should be able to decode $\tilde y^{m}$ without loss. The codelength averaged over all training and  test data
is then $mC_{L}(n,\hat{P})$ within a few bits \citep{CoverBook}. Since $\hat{P}_L(y|x)$
is based on training data, we call this the \emph{learned codelength}.

A related problem to the above is universal coding of the training
data \emph{itself}. In this case we assume the decoder knows the features
$x^{n}$ in the training data but not the corresponding labels $y^{n}$; the task
is to communicate these to the decoder. Again, we want to find a good
estimated probability law $\hat{P}_U(y|x)$ and consider the \emph{universal
codelength}
\begin{equation}
	C_{U}(n)=E_{x^n,y^n}\left[-\log\hat{P}_U(y^{n}|x^{n})\right]
	\label{eq:CU}
\end{equation}
The expectation here is over the training data only. 
%Notice that in this case, as opposed to \emph{learned codelength}, the decoder does not know 
%which $\hat{P}_U(y|x)$ the encoder is using, and some bits are therefore needed to encode this information, either explicitly or implicitly. This is a key difference
%with learned coding, see Fig.\ref{fig:Coding}.
%\begin{figure}
%	\includegraphics[width=3.2in]{ICMLfigx}
%	\caption{Universal and learned coding. In learned coding
%	the decoder knows the coding distribution $\hat P$. Here
%	$C_{U}^{+}(n,\theta)$ and $mC_{L}^{+}(n,\theta)$ are
%	the number of bits transmitted from encoder to decoder. Dashed lines indicate side-information.}
%	\label{fig:Coding}
%\end{figure}
In many cases, universal source coding can be implemented through \emph{sequential coding} \citep{CoverBook}.
In that case, the encoder uses a distribution
$\hat P_U(y_m|y^{m-1},x^m)$, updated sequentially from $y^{m-1}$, to encode $y_m$. The decoder,
having decoded $y^{m-1}$, can also determine $\hat P_U(y_m|y^{m-1},x^m)$, and therefore decode $y_m$.
We define the sequential universal codelength
\begin{equation}
	C_{U,s}(m|m-1)=E_{x^m,y^m}\left[-\log\hat P_U(y_m|y^{m-1},x^m)\right] \label{eq:CUsm}
\end{equation}
This gives a total codelength
\begin{equation}
	C_{U,s}(n) = \sum_{m=1}^n C_{U,s}(m|m-1) \label{eq:CUs}
\end{equation}
In many cases sequential and block coding
give almost the same codelength \citep{CoverBook,Shamir06},
and therefore $C_{U,s}(n) \approx C_U(n)$.

The aim of universal coding is to minimize (\ref{eq:CU}) or (\ref{eq:CUs}) according to some criterion (e.g., minimax
redundancy \citep{Shamir06}). In (\ref{eq:CUs}) the minimization
can be done term by term, by minimizing (\ref{eq:CUsm}). We notice that this
is equivalent to (\ref{CL1.eq}), and we therefore have
\begin{align}
  C_L(n) &= C_{U,s}(n+1)-C_{U,s}(n) \approx C_{U}(n+1)-C_{U}(n) \label{eq:CLapprox}
\end{align}
We clearly also have $C_L(n) \approx  C_{U,s}(n)-C_{U,s}(n-1)$ without much error; the latter is a function of training data only. Thus, we can use
(\ref{eq:CLapprox}) for finding the learned codelength (i.e., the generalization error) using universal source coding of the training data itself.

%For use in for example model selection, w
We are generally interested
in the generalization error for a specific set of training data
\citep{HastieBook}, which we can write as
\begin{equation}
C_{L}(n|x^n,y^n)=E_{x_0,y_0}\left[\left.-\log\hat{P}_L(y|x)\right|x^n,y^n\right]
\end{equation}
Notice that the expectation here is over only $(y_0,x_0)$, while the training
$(x^n,y^n)$ is fixed. Corresponding to this we have the universal
codelength for a \emph{specific} training sequence 
$C_U(n,(x^n,y^n))\approx C_{U,s}(n,(x^n,y^n))$.
However, we no longer have $C_L(n|x^n,y^n) \approx  C_{U,s}(n,(x^n,y^n))-C_{U,s}(n-1,(x^{n-1},y^{n-1}))$, since
the right hand side is calculated for a single sample $x_n$ and is not an expectation.
Instead we propose the following estimate
\begin{align}
\boxed{
  \hat C_L(n|x^n,y^n) = \frac 1 {n-m}\left( C_{U,s}(n,(x^n,y^n))-C_{U,s}(m,(x^{m},t^{m}))\right)}  \label{eq:DDL}
\end{align}

for some $m < n$; we will discuss how to choose $m$ later. We call this \emph{differential description length (DDL)} as a difference
between two description lengths.

\begin{table}
\hrule
\begin{enumerate}
  \item $C_{U}(n,(x^n,y^n))-C_{U}(m,(x^{m},y^{m}))$, the codelength difference of two block coders.
  \item $C_{U}(n,(x^n,y^n)|x^m,y^m)$, the codelength to encode $(x^n,y^n)$ when the decoder knows $(x^m,y^m)$.
  \item $C_{U,s}(n,(x^n,y^n)|m)$ the codelength to sequentially encode $(x^n,y^n)$ starting from sample $m$.
\end{enumerate}
\hrule
\caption{\label{tab:DDLmethods}DDL methods.}
\end{table}
There are three distinct ways we can calculate DDL, see Table \ref{tab:DDLmethods}. The first method might be the fastest. It can be implemented by either
finding simple block coders, or even using explicit expressions for codelength from the coding literature. The last method might
be the most general as sequential coders can be found for many problems,
and these can be nearly as fast a block coders. The second method is
attractive, as it avoid the difference of codelength in the first method, but this is not a standard coding problem. Table \ref{tab:DDLmethods} should become more clear when we discuss
a concrete example below.

In most cases, these three methods will give nearly the same
result, so the choice is more a question about finding suitable
methodology and of complexity. One of the advantages of using coding is exactly the equivalence of these methodologies.

%\begin{thm}[May not have time to finish]
%	DDL is always a strictly better estimator of generalization error than cross-validation.
%\end{thm}

\subsection{Analysis}
We will consider a simple problem that allows for analysis. The observations $x$ is from a finite alphabet with $K$ symbols,
while the labels $y$ are binary. 
The unknown parameter are the $K$ conditional distributions
$P(1|x) \doteq P(y=1|x), x\in \{1,\ldots,K\}$. It is clear that for coding (and
estimation) this can be divided into $K$ independent substreams corresponding to
$x_i \in \{1,\ldots,K\}$. Each substream can then be coded as
\citet[Section 13.2]{CoverBook}. We will demonstrate the 3 procedures
in Table \ref{tab:DDLmethods} for this  problem.
For the first procedure we use a block coder  as follows. The encoder counts the number
$k_x$ of ones in $y^n$ for the samples where $x_i=x$. It transmits $k_x$ using $n_x$ bits (since the
decoder knows $x^n$ it also knows the number $n_x$), and
then which sequences with $k_x$ bits was seen using 
$\log \left(\begin{matrix}
  k_x \\
  n_x
\end{matrix}
\right)$ bits. The codelengths
can in fact be calculated quite accurately from \citet{Shamir06}
\begin{align}
  C_U(n,(x^n,y^n)) &= \sum_{x=1}^K n_x H\left(\frac {k_x}{n_x}\right) 
  +\frac 1 2\log \frac{n_x}2 + \epsilon\left(\frac 1 n\right) \label{eq:CUbinary}
\end{align}
where $H(\cdot)$ is entropy and the term $\epsilon\left(\frac 1 n\right)\to 0$ as $n\to\infty$. This expression can now be used directly for procedure 1
in Table \ref{tab:DDLmethods}.

For procedure 3 in Table \ref{tab:DDLmethods} we can use a sequential
coder with the KT-estimator \citep{CoverBook,KrichevskyTrofimov81}
\begin{align}
	\hat P_m(1|x) &= \frac{k_x(m-1)+\frac 1 2}{n_x(m-1)+1},
\end{align}
where $k_x(m-1)$ is the number of ones corresponding to $x$ seen in $(x^{m-1},y^{m-1})$. The estimate $\hat P_m(1|x)$ is then used to encode
$y_m$. The resulting codelength is the same as (\ref{eq:CUbinary}).

Procedure 2 is not a standard coding problem. We have to encode the
number of ones  seen in the sequence $y_{m+1},\ldots,y_n$ corresponding
to the observation $x$. There are between $0$ and $n_x(n)-n_x(m)$ ones. However,
since we have already observed the number of ones $k_x(m)$ in the sequence $y^m$ this
should be used for coding. We can use Bayes rule
\begin{align}
	P(\theta|k_x(m)) &= \frac{P(k_x(m)|\theta)P(\theta)}{P(k_x(m))} = 
	(m+1) \left(\begin{matrix}
  k_x(m) \\
  n_x(m)
\end{matrix}\right)\theta^{k_x(m)}(1-\theta)^{n_x(m)-k_x(m)}.
\end{align}
which, after quantization, can be used to encode the number of ones. We do not know
if this gives the same codelength as the two previous procedures; we will leave
this procedure for a later paper.

We can use the expression (\ref{eq:CUbinary}) to analyze the performance
of DDL as a predictor of generalization error. We can rewrite
\begin{align}
  \frac {C_U(n,(x^n,y^n))} n & = \sum_{x=1}^K\hat P(x) H\left(\hat P(1|x)\right) 
  +\frac 1 {2n}\log \frac{\hat P(x)}2 + \frac K {2n}\log n + o\left(\frac 1 n\right) \nonumber\\
  & \approx H(Y|X)+\sum_{x=1}^K \left[(\hat P(x)-P(x))H(P(1|x)) \vphantom{\frac {P(x)}{2}} \right. \nonumber\\
  &\left.+P(x)\log\left(P(1|x)^{-1}-1\right)(\hat P(1|x)-P(1|x)) + \frac 1 {2n}\log \frac {P(x)}{2}\right]\nonumber\\
  & +\frac K {2n}\log n + o\left(\frac 1 n\right) \label{eq:CUexp}
\end{align}
The actual generalization error is
\begin{align}
  G &= H(Y|X)+D(P\|\hat P) = \sum_{x=1}^K H(Y|x)P(x) + D(P(Y|x)\|\hat P(Y|x))P(x) \\
  & = H(Y|X)+\sum_{x=1}^K \frac{(\hat P(1|x)-P(1|x))^2}
  {P(1|x)(1-P(1|x))\ln 4}P(x) + o\left(\frac 1 n\right)
\end{align}
The DDL estimate $\hat G$ of generalization error is given by inserting the expression
(\ref{eq:CUexp}) in (\ref{eq:DDL}).
A straightforward calculation shows that the error in this estimate is (ignoring $o(\cdot)$ terms)
\begin{align}
	\hat G - G &= \frac{K}{2(n-m)}\log \frac n m  + \sum_{x=1}^K (\hat P_n(x)-\hat P_m(x))H(P(1|x)) \nonumber \\
	& + \sum_{x=1}^K P(x)\log\left(P(1|x)^{-1}-1\right)(\hat P_n(1|x)-\hat P_m(1|x)) \nonumber\\
	& + \sum_{x=1}^K \frac{(\hat P_n(1|x)-P(1|x))^2}
  {P(1|x)(1-P(1|x))\ln 4}P(x) \label{eq:Gerr}
\end{align}
This expression can be used to analyze the performance of DDL. 
One major question for performance is how to choose $m$ in DDL (\ref{eq:DDL}).
We approach this by minimizing $E[(\hat G -G)^2]$ with respect to $m$. The
last term in (\ref{eq:Gerr}) does not depend on $m$. The second and third have zero
mean. The second term has variance
\begin{align}
	\sum_{x=1}^K\mathrm{var}\left[(\hat P_n(x)-\hat P_m(x))H(P(1|x))\right]
	&= \sum_{x=1}^K P(x)(1-P(x))\frac{1}{n-m} H(P(1|x))^2 \nonumber \\
	& \leq \frac{1}{n-m}
\end{align}
while the third term gives a variance
\begin{align}
	\lefteqn{\sum_{x=1}^K \mathrm{var}\left[P(x)\log\left(P(1|x)^{-1}-1\right)(\hat P_n(1|x)-\hat P_m(1|x))\right]} \nonumber \\
	& = \frac 1 {n-m}\sum_{x=1}^K P(x)^2 \log\left(P(1|x)^{-1}-1\right)^2P(1|x)(1-P(1|x)) 
	 \leq \frac 1 {n-m}
\end{align}
since the function $p(1-p)\log(p^{-1}-1)\leq 1$.
In conclusion we can write
\begin{align}
	E[(\hat G -G)^2] &= \left(\frac{K}{2(n-m)}\log \frac n m\right)^2
	+\frac{f(K)}{n-m}+ \frac{K}{2n(n-m)}\log \frac n m + \text{const.}\nonumber \\
	& = \frac{K^2}{2n^2 (1-\alpha)^2}\log ^2 \alpha 
	+\frac{f(K)}{n(1-\alpha)} - \frac{K}{2n^2(1-\alpha)}\log  \alpha+ \text{const.}
	\label{eq:MSE}
\end{align}
where $\alpha=\frac m n$ and $f(K)\leq 2$. We can minimize this with respect to $\alpha$ by taking derivatives
\begin{align}
	\lefteqn{\frac{\partial E[(\hat G -G)^2]}{\partial \alpha}} \nonumber \\ &=
	\frac{(\alpha-1) \ln 2 (\alpha f(K) n \ln 4+(\alpha-1) K)-2 \alpha K^2 \ln ^2 \alpha+(\alpha-1) K \ln \alpha (2 K-\alpha \ln 2)}{2 (\alpha-1)^3 \alpha n^2 \ln ^2 2}
\end{align}
Then $\frac{\partial E[(\hat G -G)^2]}{\partial \alpha}=0$ gives
\begin{align}
	  & \!\!\!\!\!\!\!\!\!\!\!\!\!\!\!\!\!\!\!\!\!\!\!\!\!\!n = \frac{K (-\alpha+\alpha \ln \alpha+1) (2 K \ln \alpha+(\alpha-1) \ln 2)}{(\alpha-1) \alpha f(K) \ln 2 \ln 4} \nonumber \\
	& \!\!\!\!\!\!\!\!\!\!\!\!\!\!\!\!\!\!\!\!\!\!\sim -\frac{2K^2  \ln \alpha}{\alpha f(K) \ln 2 \ln 4}
\end{align}
where the last expression is found by keeping dominating terms for $n$ large. 
\ifthenelse{\boolean{submit}}
{
\begin{wrapfigure}[14]{r}{3.2in}
\vspace{-0.2in}
	\begin{centering}
  \includegraphics[width=3.3in]{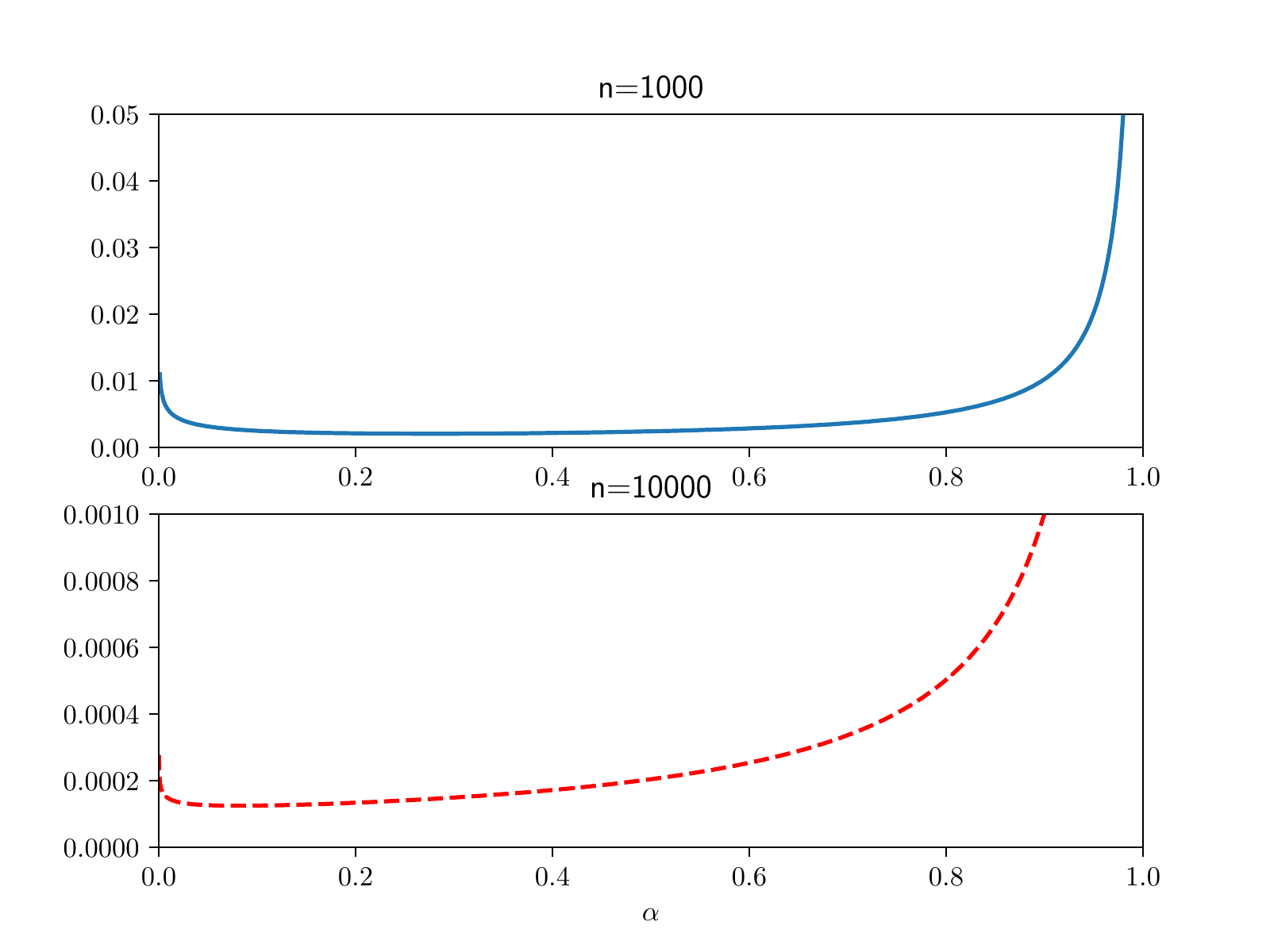}
  \vspace{-0.3in}
  \caption{\label{fig:DDL_error}The MSE for DDL as a function of $\alpha$ for
  $K=10$ and $f(K)=1$.}
 \end{centering}
\end{wrapfigure}

}{
\begin{wrapfigure}[18]{R}{3.2in}
\vspace{-0.2in}
	\begin{centering}
  \includegraphics[width=3.3in]{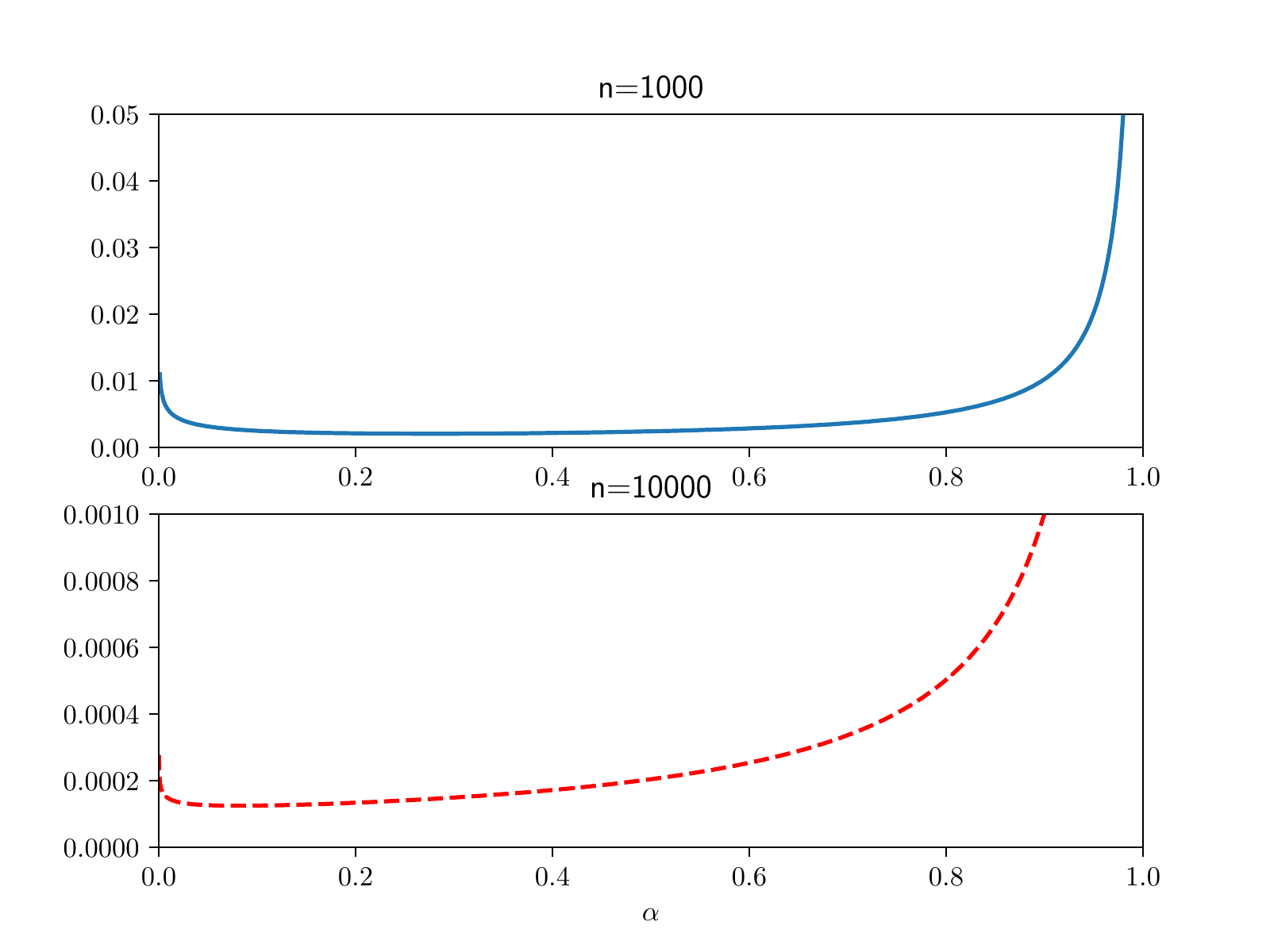}
  \vspace{-0.3in}
  \caption{\label{fig:DDL_error}The MSE for DDL as a function of $\alpha$ for
  $K=10$ and $f(K)=1$.}
 \end{centering}
\end{wrapfigure}

}
From this
\begin{align}
	\alpha & = \frac{2K^2W\left(\frac{nf(K) \ln 2 \ln 4}{2K^2}\right)}{nf(K) \ln 2 \ln 4}  \nonumber \\
	&\approx \frac{2K^2 }{f(K) \ln 2 \ln 4}\frac{\ln n}{n},
\end{align}
where $W$ is the Lambert $W$ function. 
%We can also write it as
%$m \approx \frac{2K^2 }{f(K) \ln 2 \ln 4}\ln n$.

This gives us some insight into how to choose $m$ and how DDL works. The optimum value of $\alpha$ converges
to zero as $n\to\infty$. This might be somewhat surprising. In cross-validation,
usually a fixed percentage of data is used for validation. If one thinks
of the data $(x_{m+1},y_{m+1}),\ldots,(x_n,y_n)$ as `validation' data in 
DDL, almost the entire training set is used for validation.
Of course, ordinary MDL corresponds to $\alpha=0$, so
in some sense this validates using MDL for machine learning. Yet, $\alpha=0$ gives
an infinite error, so the take away is that the gain from DDL is to avoid
this singularity. Another insight is that the optimum value of $\alpha$ increases as
$K^2$ with the number of parameters. The factor $f(K)$ is less predictable, but
at least we know it is bounded as a function of $K$. Thus, complex models require
a  large value of $\alpha$.

In general it of course can be difficult to find the exact optimum value of
$\alpha$. However, the function in (\ref{eq:MSE}) is quite insensitive to
$\alpha$. This is most easily seen numerically, see Fig. \ref{fig:DDL_error}.
There is a large plateau from perhaps 0.1 to 0.7 where the error is close
to the minimum.

%\begin{figure}[hbt]
%\begin{centering}
%  \includegraphics[width=3.5in]{DDL_error}
%  \caption{\label{fig:DDL_error}The MSE for DDL as a function of $\alpha$ for
%  $K=10$ and $f(K)=1$.}
% \end{centering}
%\end{figure}

\subsection{Use for model selection}
\label{sec:ModelSelection}
\begin{wrapfigure}[17]{R}{3.2in}
	\begin{centering}
	\vspace{-0.7in}
  \includegraphics[width=3.3in]{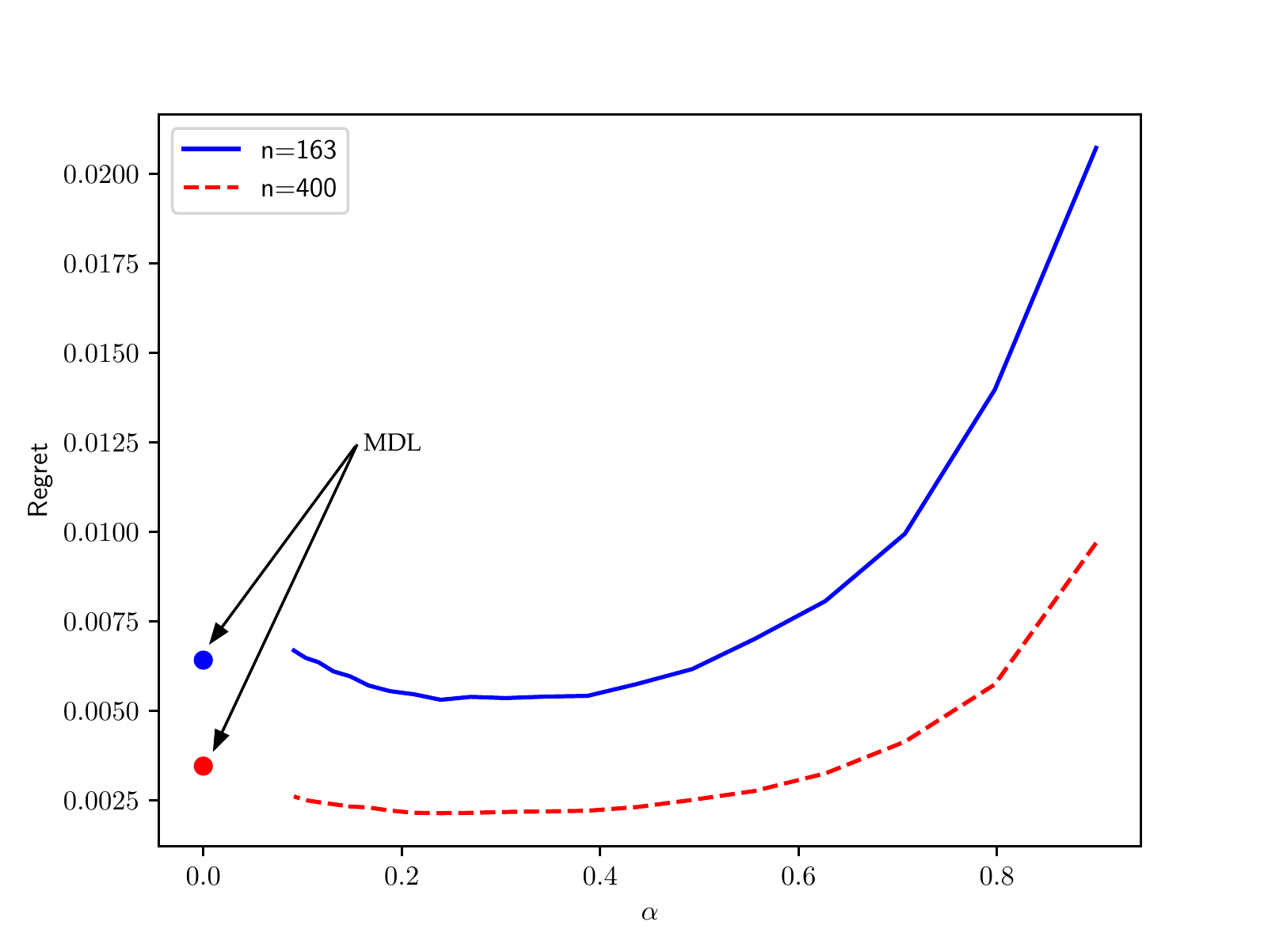}
  \vspace{-0.3in}
  \caption{\label{Regretnt.fig}The figure shows generalization error versus
$n_{t}$ when data is dependent or independent for different values
of $n$. The dots are the generalization error for full description
length (MDL).}
 \end{centering}
\end{wrapfigure}
We will illustrate how the above methodology works for model selection
for a simple example.
The data $(x,y)$ is (iid) binary given by a conditional probability
distribution $P(y|x)$ (and marginal $P_{x}(x)$). Under model $\mathcal{M}_{1}$,
$y$ is independent of $x$, while under $\mathcal{M}_{2}$, $y$ is
dependent on $x$. For model $\mathcal{M}_{1}$ there is a single
unknown parameter $\theta_{1}=p(1|0)=p(1|1)$, while for $\mathcal{M}_{2}$
there are two unknown parameters $\theta_{2}=(p(1|0),p(1|1))$. 

Fig. \ref{Regretnt.fig} shows the worst case regret (\ref{eq:regret}) as a function of $\alpha$. The worst case  regret is found by maximizing (numerically) the regret over the parameters $\theta$, $P_{x}(x)$, and the training data.
The main point here, as indicated above, is that the regret is fairly
insensitive to $\alpha$, and for a wide range of values of $\alpha$ the
regret from DDL is smaller than MDL.
%\begin{figure}[tbh]
%\begin{centering}
%\vspace{-0.13in}
%\includegraphics[width=3.5in]{BinaryGEn}
%\par\end{centering}
%\vspace{-0.2in}
%\caption{\label{Regretn.fig}The figure shows worst case generalization error
%when data is dependent or independent, for either differential description
%length with optimum $n_{t}$ or full description length.}
%% These figures generated by testgenerrbin with data
%% in testgenerrbin5.dill
%\end{figure}
%The remaining issue is how to choose $n_{t}$. Fig. \ref{Regretnt.fig}
%shows the regret as a function of $n_{t}$. The main conclusion is
%that unless we choose $n_{t}$ very small or very large, the exact
%value has little effect. It seems that a good simple choice could be $n_{t}=\frac{n}{2}$.

%\begin{figure}[tbh]
%\begin{centering}
%\vspace{-0.16in}
%\includegraphics[width=3.5in]{BinaryGEalpha.pdf}
%\par\end{centering}
%\vspace{-0.2in}
%\caption{\label{Regretnt.fig}The figure shows generalization error versus
%$n_{t}$ when data is dependent or independent for different values
%of $n$. The dots are the generalization error for full description
%length (MDL).}
%
%\end{figure}

\section{Hyperparameter Selection in Machine Learning}

%In general we can write the codelength to
%encode \emph{training} data as
%\begin{equation}
%C=-\log P(t^{n}|x^{n};\hat{\theta}_{h},h),\label{CL.eq}
%\end{equation}
%where $h$ denotes the hyperparameters and $\hat{\theta}_{h}$ is the maximum likelihood estimate.
%This is a codelength, but it requires the decoder to know $\hat{\theta}_{h}$ (both
%encoder and decoder are assumed to know $h$).
We consider a machine learning problem specified by a set of parameters
$\theta_h$ that are estimated from the training data, and a set of hyperparameters
$h$ that are chosen; the aim is to choose $h$ to minimize the generalization
error (\ref{eq:generror}) for log-loss (cross-entropy)
We focus on procedure 3 in Table \ref{tab:DDLmethods}, as this can
be easily implemented for many ML methods and does give an actual
codelength, not just an approximation. 
Specifically, we calculate the codelength through
\begin{equation}
C_{U,s}(n,(x^n,y^n)|m)=-\sum_{i=m}^{n-1}\log P(y_{i+1}|x_{i+1};\hat{\theta}_{h}(y^{i},x^{i}),h),\label{predictiveMDL.eq}
\end{equation}
where $\hat{\theta}_{h}(y^{i},x^{i}) $ is the maximum likelihood estimate. When $m=0$ this is Rissanen's predictive
MDL \citep{Rissanen86}.
An issue with predictive MDL is initialization: $\hat{\theta}_{h}(t^{i},x^{i})$
is clearly not defined for $i=0$, and likely $i$ should be large
for the estimate to be good. When the initial estimate is poor,
it can lead to  long and arbitrary codelengths, see \citet{Sabeti2017ISITpredictive}.
An advantage of  DDL is that it completely overcomes this problem because 
$m>0$.
\subsection{Linear Regression}
\label{sec:LR}
We can implement DDL for linear regression directly through
(\ref{predictiveMDL.eq}). Since regression can
be implemented recursively \citep{HaykinBook}, this is
very efficient.
%Let $\boldsymbol{\Phi}_{m}=[\boldsymbol{\phi}(x_{1})\ldots\boldsymbol{\phi}(x_{m})]$,
%where $\boldsymbol{\phi}(x_{i})$ are the feature vectors. Assuming
%a Gaussian model with variance $\beta$, the ML estimate with regularization
%is (e.g., \citet{BishopBook,ScharfBook})
%\begin{align}
%\hat{\mathbf{w}}_{m} & =\left(\boldsymbol{\Phi}_{m}^{T}\boldsymbol{\Phi}_{m}+\lambda\mathbf{I}\right)^{-1}\boldsymbol{\Phi}_{m}\mathbf{t}_{m}\nonumber \\
%\hat{\beta}_{m} & =\left(\frac{1}{m}\sum_{i=1}^{m}(t_{i}-\hat{\mathbf{w}}_{m}^{T}\boldsymbol{\phi}(x_{n}))^{2}\right)^{-1}\label{LS.eq}
%\end{align}
%%and the predictive MDL (\ref{predictiveMDL.eq}) is explicitly
%%\begin{align*}
%%C & =\sum_{i=n-n_{t}-1}^{n}-\log f(t_{i+1};\hat{\mathbf{w}}_{i},\hat{\beta}_{i})\\
%%f(t_{i+1};\hat{\mathbf{w}}_{i},\hat{\beta}_{i}) & =\frac{1}{\sqrt{2\pi\hat{\beta}_{i}}}\exp\left(-\frac{(t_{i+1}-\hat{\mathbf{w}}_{i}^{T}\boldsymbol{\phi}(\mathbf{x}_{i+1})^{2}}{2\hat{\beta}_{i}}\right)
%%\end{align*}
%The estimate (\ref{LS.eq}) is not defined until $m$ is at least
%equal to the dimension of the feature space. But even then, the estimate
%is not reliable, and using this directly for MDL can give a codelength
%which is nearly infinite, which makes predictive MDL not very useful.
%However, with DDL  we only need to calculate (\ref{LS.eq})
%for $m\geq n-n_{t}$, which makes it much more reliable. There are
%recursive algorithms for updating $\hat{\mathbf{w}}_{m}$ \citep{HaykinBook},
%so predictive MDL/DDL can be implemented very efficiently.

\begin{figure}[tbh]
\vspace{-0.15in}
\hspace{-0.2in}
\includegraphics[width=3.1in]{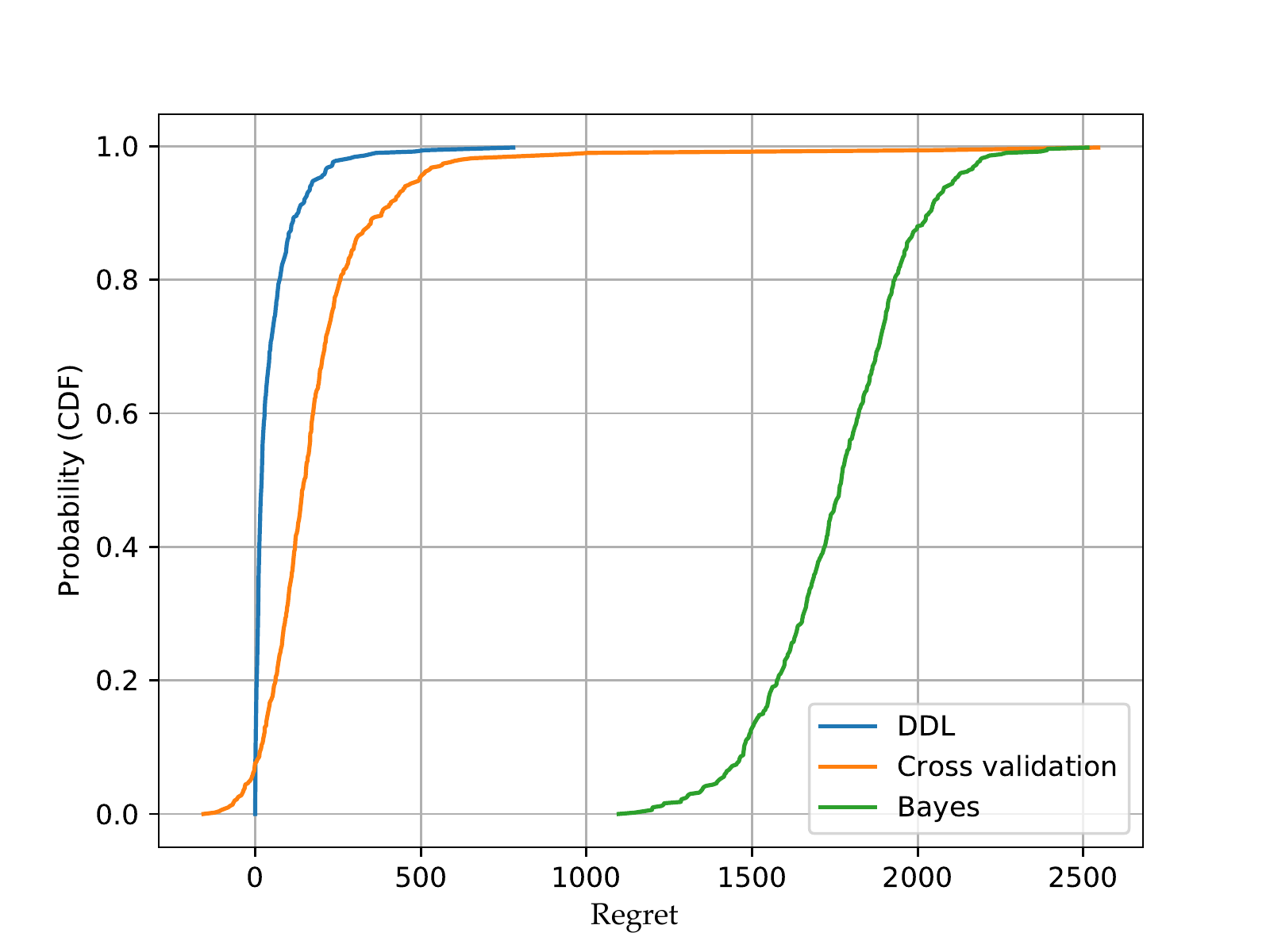}
\hspace{-0.25in}
\includegraphics[width=3.1in]{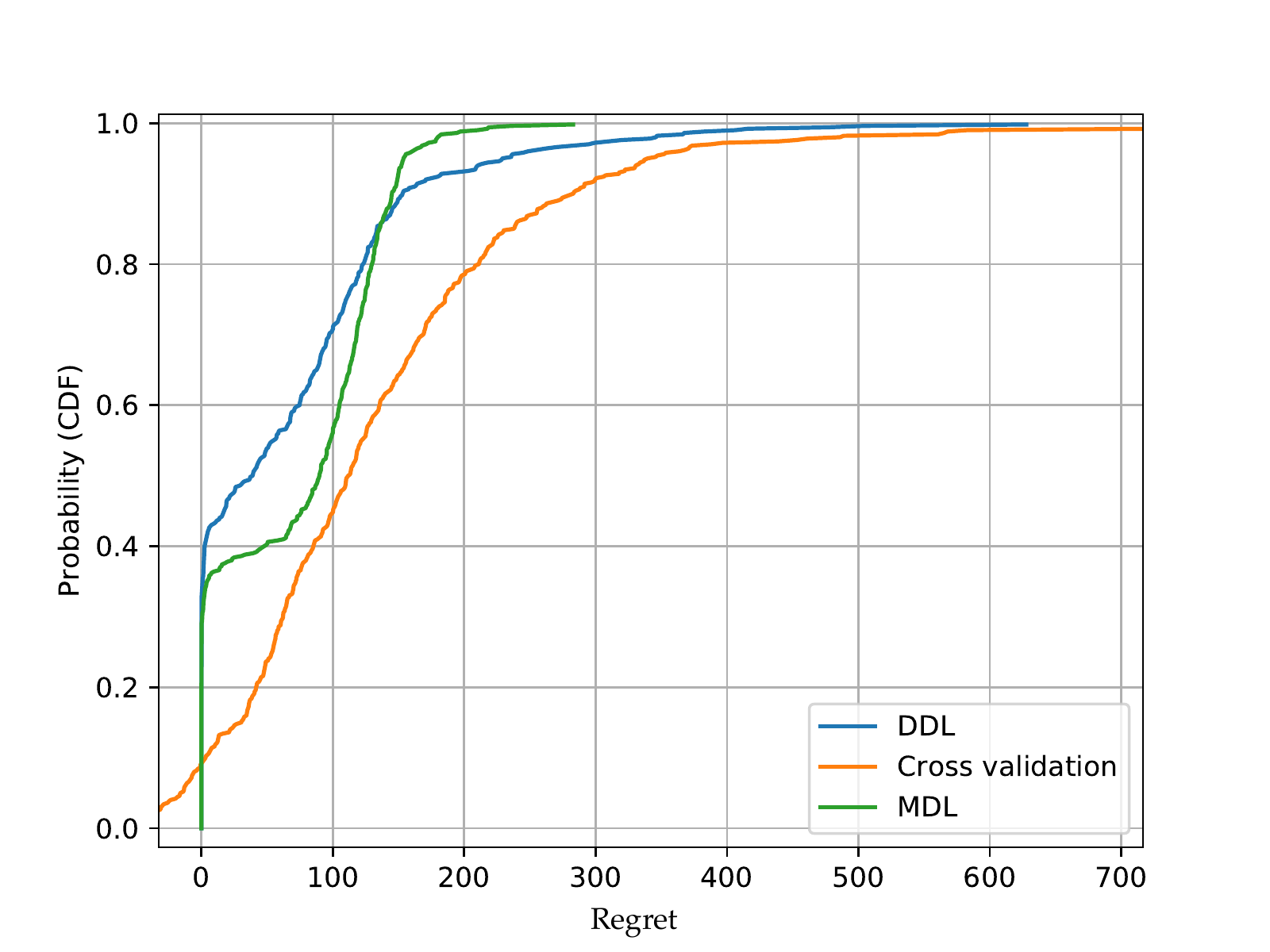}
\vspace{-0.25in}
\center{\hspace*{0.5in}(a) \hspace*{3in} (b)}

\vspace{-0.1in}
\caption{\label{LS.fig}\label{fig:LS1order}Generalization error for least squares for simple curve
fitting. The curves show the distribution
of regret in terms of the CDF. (a) is for
regularized LS and (b) for order selection.}
\vspace{-0.2in}
\end{figure}

Figure \ref{LS.fig} shows some experimental results. The setup is
that of fitting polynomials of order up to 20 to the curve $\sin(3x),x\in[-2,2]$.
We generate 500 random $x_{n}$ and observe $y_{n}=\sin(3x_{n})+w_{n}$,
where $w_{n}\sim\mathcal{N}(0,0.15)$. We seek to optimize the regularization
parameter $\lambda$ in $L_2$ regularization. We use DDL with $m=\frac{n}{2}$ and compare
with cross-validation, where we use 25\% of samples for cross-validation.
We also compare with Bayes model selection, using the theory in \citet[Section 3.5.1]{BishopBook}
to optimize $\alpha=\lambda\hat{\beta}$ (notation from \citet[Section 3.5.1]{BishopBook}). We plot the regret
(\ref{eq:regret}). 
One can see that DDL  chooses the
correct $\lambda$ in nearly 50\% of cases, and is always better than
cross-validation
(The reason cross-validation can have
negative regret is that $\hat{w}$ is calculated from only 75\% of samples,
and that has a chance of being better than an estimate calculated
from the full set of training samples). 
It is also better than the Bayes method (which, in its defense,
was not developed specifically to minimize generalization error).
%The curves for MSE rather than log-loss are nearly identical, so we
%have not included them.

In Fig. \ref{fig:LS1order} we modify the experiment to directly
varying the model order $M$ without regularization. 
In that case, we can also compare with traditional MDL \citep{Rissanen83} through
the  approximation \citep{Rissanen83}%of (\ref{MDL1.eq}) by
\begin{equation*}
C_{U}(n,(x^n,y^n))=\min_{\hat{\theta}_{h}}-\log P(y^{n}|x^{n};\hat{\theta}_{h},h)+\frac M 2 \log n
\end{equation*}
We see that DDL is again better than cross-validation, and better
than traditional MDL, except that DDL and cross-validation have heavy tails.

%\begin{figure}[tbh]
%\vspace{-0.1in}
%\includegraphics[width=3.5in]{LS1order}
%%\vspace{-0.3in}
%\caption{Generalization error for least squares for simple curve fitting without
%regularization with order optimization. The curves show the distribution
%of regret error in terms of the CDF. }
%
%\end{figure}

%\subsection{Logistic Regression}
%
%The predictive MDL (and hence the related DDL)
%% and direct quantization methods 
%can be easily generalized
%to logistic regression. We will outline this for binary labels. Rather
%than the Gaussian codelength, the codelength of the labels is now
%given by the cross-entropy: $-\sum_{i=1}^{N}t_{i}\log y_{i}+(1-t_{i})\log(1-y_{i})$,
%where $y_{i}=\sigma(\mathbf{w}^{T}\mathbf{x}_{i})$ with
%$\sigma$ the sigmoid function \citep{BishopBook}. For the predictive
%MDL one needs to calculate the maximum likelihood solution $\hat{\mathbf{w}}_{m}$
%for $n-n_{t}<m\leq n$, which can be done with iterative reweighted
%least squares (IRLS) \citep{BishopBook}. The IRLS is iterative. Since
%$\hat{\mathbf{w}}_{m+1}\approx\hat{\mathbf{w}}_{m}$, one can initialize
%the iteration for $\hat{\mathbf{w}}_{m+1}$ with $\hat{\mathbf{w}}_{m}$,
%and only a few iterations are needed. Calculating all the $\hat{\mathbf{w}}_{m}$
%therefore is not prohibitively complex. 

\subsection{Neural Networks}
MDL and coding theory is based on maximum likelihood solutions e.g. in (\ref{predictiveMDL.eq}).
%, i.e., (\ref{MDL1.eq}). 
On the other
hand, training of a neural network is unlikely to converge to the
maximum likelihood solution. Rather, the error function has many local
minima, and training generally iterates to some local minimum (or
a point near a local minimum), and which one can depend on the initialization
condition. %, i.e., a solution of the type in (\ref{eq:localMDL}). 
This requires adaption of methods like  (\ref{predictiveMDL.eq}). Another challenge
is complexity. For example, directly using predictive MDL (\ref{predictiveMDL.eq})
requires training for every subset of samples $(x^{i},y^{i})$ for
$i=m+1,\ldots,n$, which is not computationally feasible. 
%Applying
%description length to neural networks in a meaningful and practical
%way therefore is highly non-trivial. 

%There are many methods for training neural networks. Our aim is \emph{not} to develop new training methods, but rather to use
%description length for hyperparameter optimization with any training
%method.
%We would therefore like to find the description length for a neural network with
%a \emph{specific} solution for the weights $\mathbf{w}$, fairly agnostically
%to how that solution was found. 

Our methodology is as follows. We start with a solution $w(n)$ found
by training on $(x^n, y^n)$; in order to implement DDL we first need
another solution $w(m)$ found
by training on $(x^m, y^m)$. We would like this solution to be ``related''
to $w(n)$. 
%This is in particular important if we use procedures 1 and 2 in
%Table \ref{tab:DDLmethods}, as otherwise the difference could be
%quite inaccurate. 
We therefore train on $(x^m, y^m)$ with $w(n)$ as
initialization. The idea is that the solution for $w(m)$ will be at local 
minimum near $w(n)$. However, it is important that the solution $w(m)$ is independent of $((x_{m+1},y_{m+1}),\ldots,(x_n,y_n))$; if not,
the estimate of the generalization error will actually just be the training error.
% -- this is related to forgetting in deep learning [??].
We next add the data $((x_{m+1},y_{m+1}),\ldots,(x_n,y_n))$ back sequentially while retraining, in an implementation of (\ref{predictiveMDL.eq}); this might be done in blocks rather than for individual data points, both for reasons of efficiency and because deep learning algorithms train on data in batches. This still results in a valid coder.
The idea is that we would like the sequence of solutions $w(m),w(m+1),\ldots,w(n-1)$ to converge to the original solution $w(n)$.

\begin{figure}[tbh]
	\begin{centering}
	%\vspace{-0.1in}
  \includegraphics[width=5 in]{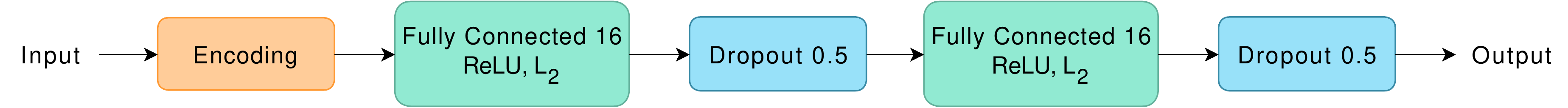}
  %\vspace{-0.1in}
  \caption{\label{fig:NN_arch} Architecture of employed neural network.}
  %\vspace{-0.1in}
 \end{centering}
\end{figure}

We test our methodology on the IMDB dataset, a real-world binary classification example, with the goal of choosing the best value for the regularization parameter $\lambda$ in $L_2$ regularization. This dataset contains 50,000 movie reviews which are split equally into training and test data, and each sample is labeled either as positive or negative \citep{maas2011learning}. A multi-layer neural network is used to learn the data as illustrated in Fig. \ref{fig:NN_arch}. The encoding module is a multi-hot encoder that converts the input sequences of words, restricted to top 5,000 most frequently occurring words, into a vector of 0s and 1s. Each of fully connected layers are made up of 16 neurons with ReLU activation function followed by the dropout layer with the rate of 0.5. Finally, sigmoid function is applied to map output into $[0,1]$ range. 

Here we compare DDL and cross-validation, where we hold out 20\% of samples for cross-validation. First we do full training over $(x^n, t^n)$ using Adam optimizer with  parameters $\alpha = 0.001, \: \beta_1 = 0.9, \: \beta_2 = 0.999, \: \text{and} \: \epsilon = 10^{-8} $ \citep{kingma2014adam}. In addition, the number of total epochs and batch size are set to 50 and 128 respectively. For training over $(x^m, t^m)$, where $m=0.8 n$, we initialize the network with weights from $(x^n, t^n)$ and try to unlearn holdout data. Since the gradient close to the local minima is very small, we train the network for few epochs with higher learning rate to unlearn the removed data and then switch to smaller values to learn just over $(x^m, t^m)$. This approach makes unlearning faster with similar complexity as training over $(x^n, t^n)$. For this experiment, we first train 40 epochs with learning rate $\alpha=0.01$  and then  back  to the initial value $\alpha = 0.001$ for 10 epochs while keeping other parameters same as the training over $(x^n, t^n)$. We then add 5 blocks of data sequentially to retrain the network by initializing from the previous step weights. To speed up  retraining, we duplicate added block multiple times until we reach the size as big as the one of the current training set. Finally, we retrain the network for only one epoch, with same parameters for training over $(x^n, y^n)$, which is less than 50 epochs used for training over $(x^n, y^n)$ and $(x^m, y^m)$. The codelength of sequential encoding is calculated according to the third procedure in Table \ref{tab:DDLmethods}.

\begin{figure}[tbh]
\vspace{-0.15in}
\hspace{-0.2in}
\includegraphics[width=3.1in]{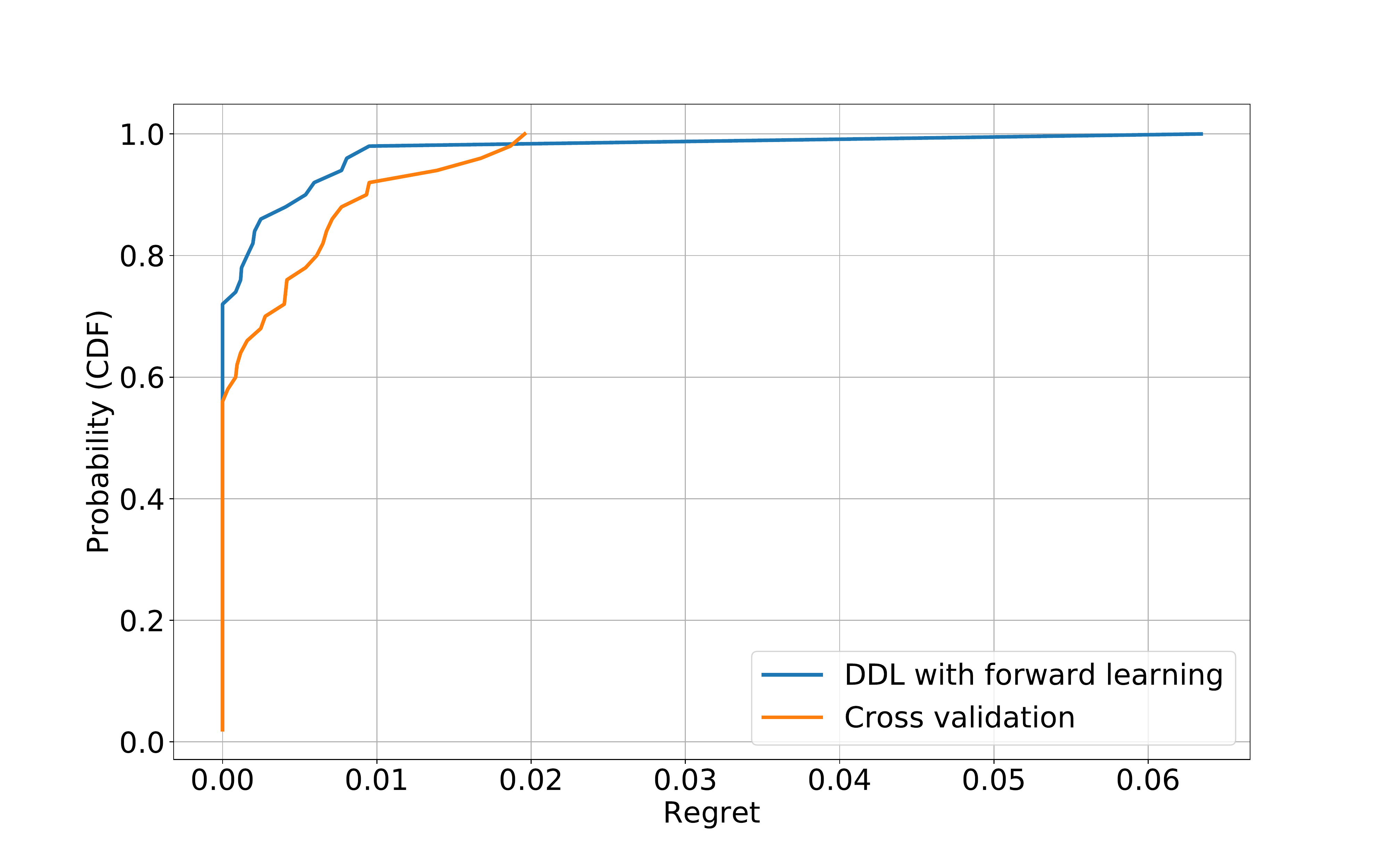}
\hspace{-0.25in}
\includegraphics[width=3.1in]{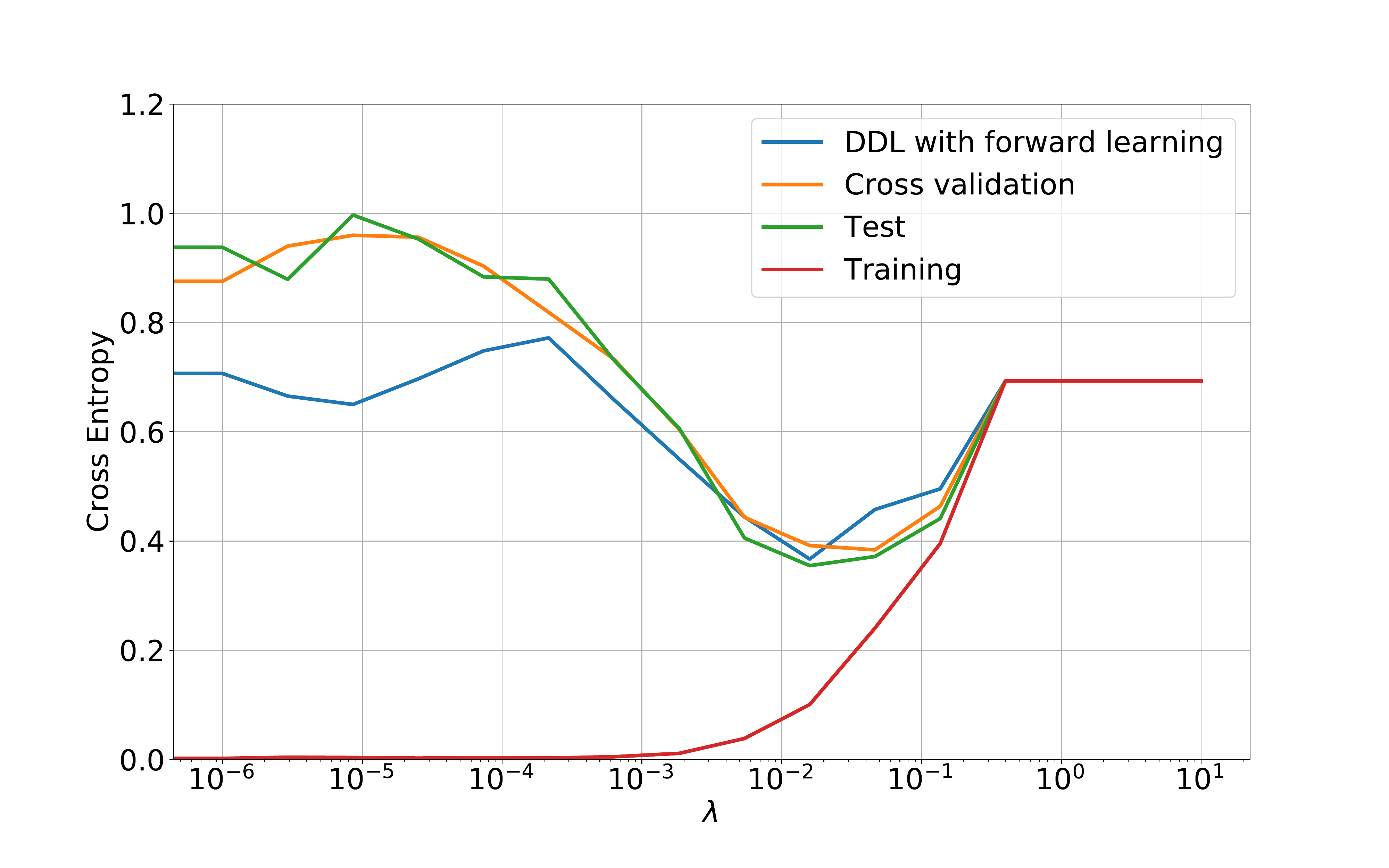}
\vspace{-0.3in}
\center{\hspace*{0.5in}(a) \hspace*{3in} (b)}

\vspace{-0.1in}
\caption{\label{fig:GE}\label{fig:CE}Generalization error for binary classification on IMDB dataset. (a) is the distribution of regret in terms of the CDF (b) is a sample of estimated generalization error by different methods vs. true generalization error.}
\vspace{-0.2in}
\end{figure}

We plot the regret (\ref{eq:regret}), as presented in Fig. \ref{fig:GE}. The results are calculated over 50 trials where, at each trial, the training set is selected randomly from the original training dataset with the size of 20\%. As it can be seen, the proposed approaches outperform cross-validation in most cases. The superiority of DDL in terms of picking the correct $\lambda$ for one typical trial is also shown in Fig. \ref{fig:CE}.

\section{Conclusion}
%In this paper we have developed algorithms for model and hyperparameter selection based on description
%length through differential description length (DDL), and showed that they work in some simple experiments.
%
This paper has developed
the framework for DDL. We will discuss this in relation to
traditional MDL. 
DDL can be used as a direct estimator of generalization error,
which MDL cannot. Both can be used for model selection, and here we can think
of DDL as a modification of MDL as follows: rather than directly using MDL for the whole dataset, we calculate the difference of
MDL for the whole dataset and a subset of the data. This difference is a better decision variable than direct MDL.
The difference can be calculated in three different ways,
as outlined in Table \ref{tab:DDLmethods}.

%In principle DDL can be used with any MDL method. However,
%because DDL is a difference it is more sensitive to accurate
%calculation of codelength. We therefore developed methods
%directly aimed at neural networks and deep learning methods.
%%TODO hyperparameter optimization algoritms refs

\newpage
\bibliographystyle{plainnat}
\bibliography{Coop06,ahmref2,Coop03,BigData,ECGandHRV}

\end{document}